\documentclass[letterpaper, 10 pt, conference]{ieeeconf}  

\IEEEoverridecommandlockouts                              

\usepackage{graphics} 
\usepackage{epsfig} 
\usepackage{amsmath} 

\title{\LARGE \bf 
Emotional EEG Classification using Upscaled Connectivity Matrices
}
\author{Chae-Won Lee$^{1}$ and Jong-Seok Lee$^{1}$
\thanks{$^{1}$The authers are with the School of Integrated Technology and the BK21 Graduate Program in Intelligent Semiconductor Technology, Yonsei University, Seoul 03722, Korea. E-mail: \{chae-won.lee, jong-seok.lee\}@yonsei.ac.kr}%
}

\begin{document}

\maketitle

\thispagestyle{empty}
\pagestyle{empty}

\begin{abstract}

Recent studies have demonstrated the effectiveness of using connectivity matrices as input to convolutional neural networks (CNNs) for emotional EEG classification, as they can effectively capture interregional interaction patterns. However, these matrices often suffer from loss of critical spatial information during convolution operations. To address this issue, we propose a simple yet effective approach: upscaling connectivity matrices to enhance local patterns. Experimental results show that this technique significantly improves classification performance, highlighting the importance of preserving spatial structures in early processing stages.
\end{abstract}

\section{Introduction}

Emotions play a vital role in shaping the user experience in modern intelligent systems, including content recommendation, virtual reality, and human-computer interaction (HCI). Consequently, there is an increasing demand for accurate and real-time emotion recognition methods. Traditional approaches based on self-reports, such as surveys and interviews, suffer from subjectivity and poor temporal resolution. In contrast, electroencephalography (EEG)-based techniques offer real-time monitoring capabilities and reduced bias, making them attractive for implicit emotion analysis \cite{moon2016implicit}.

Early EEG-based methods mainly extracted features in the amplitude or frequency domain. Although effective to some extent, these approaches often fail to capture the dynamic interactions between brain regions. To overcome this limitation, brain connectivity analysis has gained attention, focusing on the relationships among different brain regions \cite{horwitz2003elusive}, \cite{jang2018eeg}, \cite{jang2021eeg}. Connectivity matrices, constructed using metrics such as Pearson's correlation coefficient (PCC), phase locking value (PLV) and transfer entropy (TE), provide a two-dimensional representation of interregional interactions and are widely used in emotional EEG classification \cite{moon2020emotional}.

Despite promising results, convolutional neural network (CNN)-based models applied to connectivity matrices face challenges. The spatial resolution of raw connectivity matrices often limits the model’s ability to learn fine-grained features, leading to degraded classification performance. In particular, early convolution layers may overlook important local patterns, undermining the effectiveness of the model.

To address this issue, we propose an interpolation-based method that upscales connectivity matrices prior to inputing them into CNNs. This upscaling enhances the spatial granularity of the input, allowing the model to extract more discriminative features from early layers. Through extensive experiments, we show that this approach improves both classification accuracy and interpretability.

The remainder of this paper is organized as follows. Section II reviews related work. Section III describes the proposed method. Experimental setup and results are presented in Sections IV and V, respectively. Section VI concludes the paper.

\section{Related Work}

\subsection{EEG-Based Emotion Recognition}

EEG has been widely utilized in emotion recognition due to its ability to capture real-time brain activity \cite{moon2016implicit}. Both time-domain and frequency-domain analyses have been employed to interpret EEG signals associated with emotional states.

In particular, frequency-domain features have shown strong relevance to emotion classification. Liu \textit{et al.} \cite{liu2017real} distinguished emotional categories such as joy, anger, fear, and sadness using power spectral density (PSD) and asymmetry indices. Zhang \textit{et al.} \cite{zhang2016approach} further enhanced classification performance by incorporating time-frequency distribution analysis, which captures both temporal and spectral characteristics of EEG signals.

More recently, advances in deep learning have significantly improved feature extraction from EEG. Among various architectures, CNNs have demonstrated high performance in automated emotion recognition. Yanagimoto and Sugimoto \cite{yanagimoto2016convolutional} applied CNNs to classify emotional valence and arousal levels, highlighting the effectiveness of learned representations over hand-crafted features. Li \textit{et al.} \cite{li2018hierarchical} proposed transforming EEG scalp maps into 2D images for CNN processing, enabling better handling of spatial and individual variability in emotional responses.

\subsection{Brain Connectivity in Emotion Analysis}

Brain connectivity has emerged as a powerful tool in EEG-based emotion research, providing insights into functional interactions across brain regions. Unlike traditional approaches that focus on localized activation, connectivity analysis considers the dynamic relationships between distributed neural networks.

Costa \textit{et al.} \cite{costa2006eeg} demonstrated that phase synchronization patterns in EEG are significantly correlated with emotional states. Lee and Hsieh \cite{lee2014classifying} proposed a method that integrates correlation, coherence, and phase synchronization to enhance emotion classification accuracy.

The importance of connectivity-based analysis is further supported by recent neuroimaging studies. Saarimäki \textit{et al.} \cite{saarimaki2022classification} showed that functional connectivity patterns from fMRI data reliably differentiate discrete emotion categories, especially within the default mode network. In the context of EEG, a recent systematic review by Khaleghi \textit{et al.} \cite{Khaleghi2024EEG} comprehensively analyzed how various brain abnormalities alter functional connectivity patterns, providing a broader clinical perspective on network disruptions.

Building on these findings, recent works have explored the integration of brain connectivity and deep learning. Moon \textit{et al.} \cite{moon2020emotional} demonstrated that EEG-derived connectivity matrices can serve as effective inputs to CNNs for learning emotion-related features, underscoring the potential of combining neural connectivity representations with data-driven models.

\section{Proposed method}

\subsection{Connectivity Matrix Construction}

In this study, symmetric connectivity matrices are constructed using the PCC to quantify linear relationships between EEG signals. PCC evaluates the degree of synchronization between two signals and is defined as

\[
\text{PCC}(i, k) =  \frac{\frac{1}{T} \sum_{t=1}^{T}(x_i^t - \mu_i) (x_k^t - \mu_k)}{\sigma_i \sigma_k} \eqno{(1)}
\] where \( x_i^t \) and \( x_k^t \) denote EEG signals from channels \( i \) and \( k \) at time index \( t \), \( \mu_i \) and \( \mu_k \) are their respective means, and \( \sigma_i \) and \( \sigma_k \) are their standard deviations over the temporal window having length \( T \). The PCC value ranges from –1 to 1, with higher absolute values indicating stronger linear correlation.

Electrode placement within the connectivity matrix follows the distance-restricted arrangement method proposed in \cite{moon2020emotional}, which emphasizes spatial proximity among electrodes to reduce noise from long-range, spurious connections. In other words, electrodes are ordered such that those located in the same hemisphere and in close spatial proximity are positioned adjacently in the matrix. The resulting matrices are processed by CNNs to extract discriminative features for emotion classification.

\begin{figure}
\centering
    \includegraphics[width=0.5\textwidth]{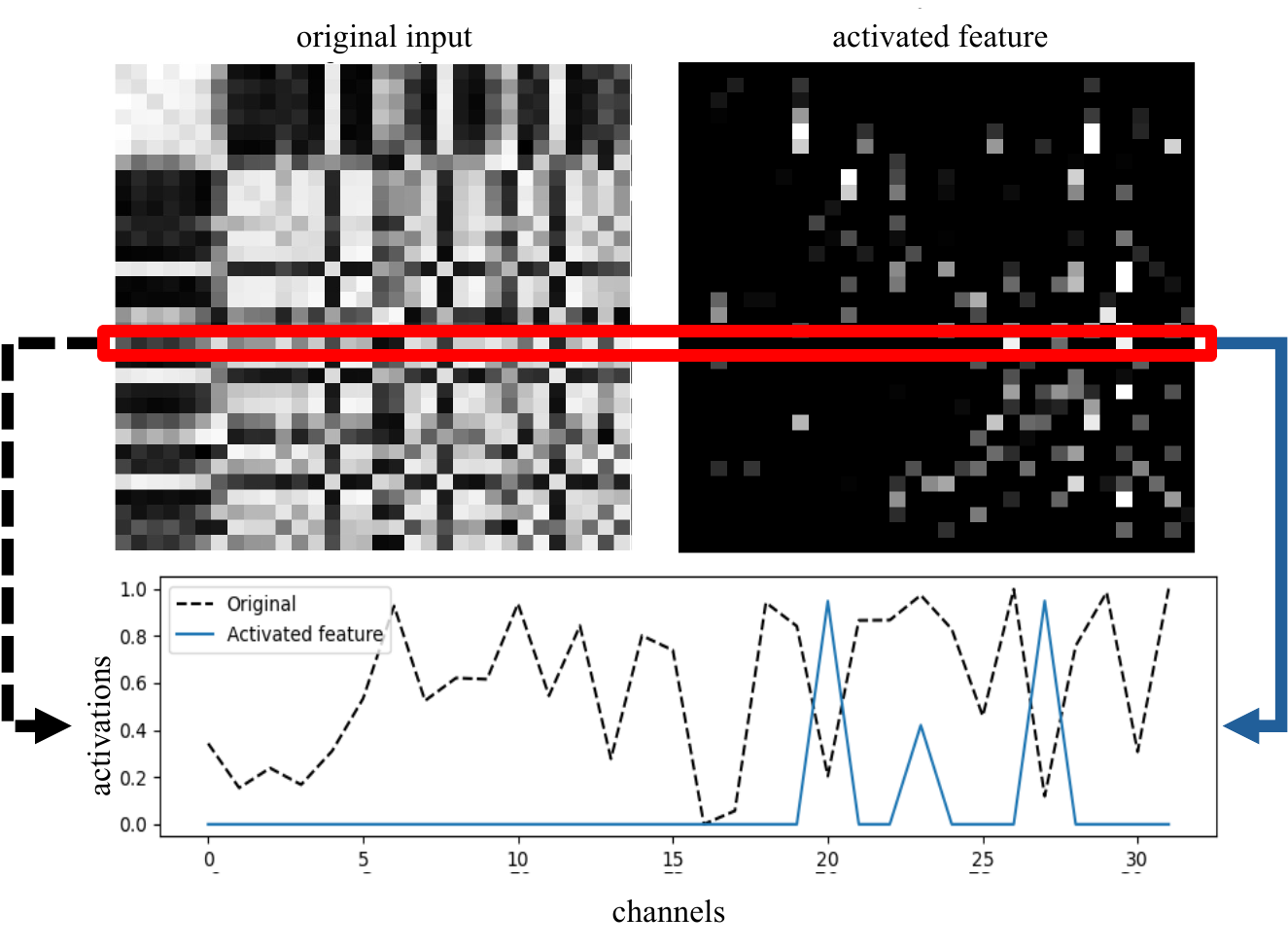}
    \caption{Activation (right) after the first convolutional layer for a connectivity matrix (left)}
    \label{fig:actconmat}
    \vspace{1cm}
    \includegraphics[width=0.5\textwidth]{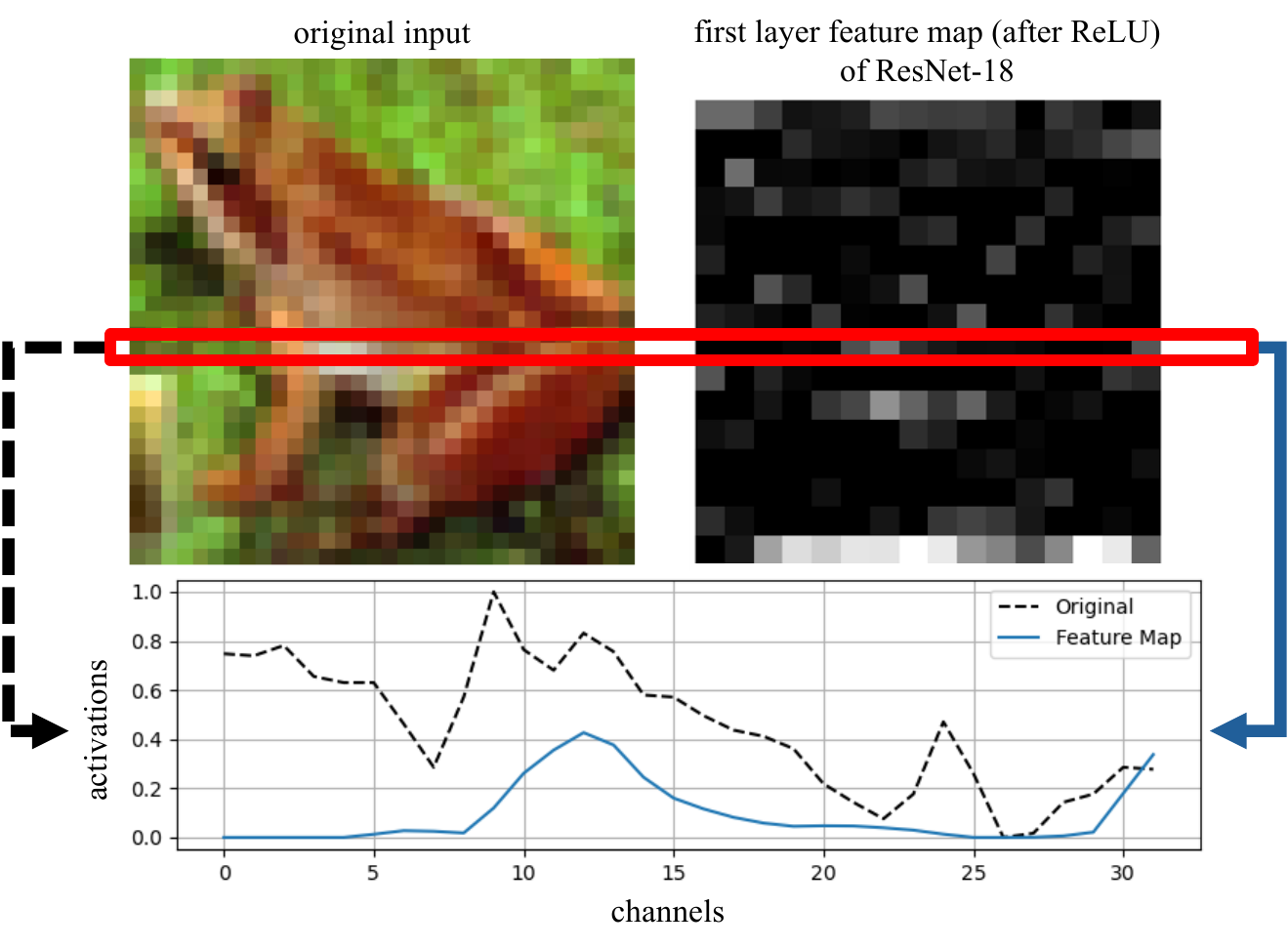}
    \caption{Activation (right) after the first convolutional layer for a natural image (left) from the CIFAR-10 dataset }
    \label{fig:actcifar}
\end{figure}
\subsection{Limitations of Conventional Approaches}

Despite their widespread adoption, CNN-based models often face challenges in effectively learning from connectivity matrices.  Fig.~\ref{fig:actconmat} compares an input connectivity matrix and its corresponding activation after the first convolutional layer, which reveals that salient local patterns in the input are not preserved well in the activation. This information loss suggests that conventional processing pipelines may inadequately capture critical relational features embedded in the connectivity matrix.

This limitation arises primarily from the intrinsic nature of connectivity matrices, which encode pairwise relationships rather than spatially continuous features. For comparison, Fig.~\ref{fig:actcifar} presents a natural image from the CIFAR-10 dataset \cite{krizhevsky2009learning} and its corresponding activation after the first convolutional layer of a CNN trained for object recognition. In this case, the model successfully captures meaningful structures such as edges and textures, as the input contains coherent spatial information. However, in the case of connectivity matrices, the absence of such spatial continuity impedes the CNN's ability to learn effective representations.

Consequently, this mismatch leads to inconsistent activation patterns, reduced classification accuracy, and diminished model interpretability. These observations underscore the necessity for preprocessing strategies that enhance the spatial structure of connectivity matrices, enabling more effective feature extraction in CNN-based models.

\begin{table}[h!]
\centering
\caption{Sub-frequency bands employed for EEG decomposition}
\begin{tabular}{|c|c||c|c|}
\hline
\textbf{Band} & \textbf{Range (Hz)} & \textbf{Band} & \textbf{Range (Hz)} \\
\hline
Delta & 0--3 & Low-beta & 13--16 \\
Theta & 4--7 & Mid-beta & 17--20 \\
Low-alpha & 8--9.5 & High-beta & 21--29 \\
High-alpha & 10.5--12 & Beta & 13--29 \\
Alpha & 8--12 & Gamma & 30--50 \\
\hline
\end{tabular}
\label{table:freq}
\end{table}

\begin{figure*}[t]
    \vspace{0.5cm}
    \centering
    \includegraphics[width=2\columnwidth]{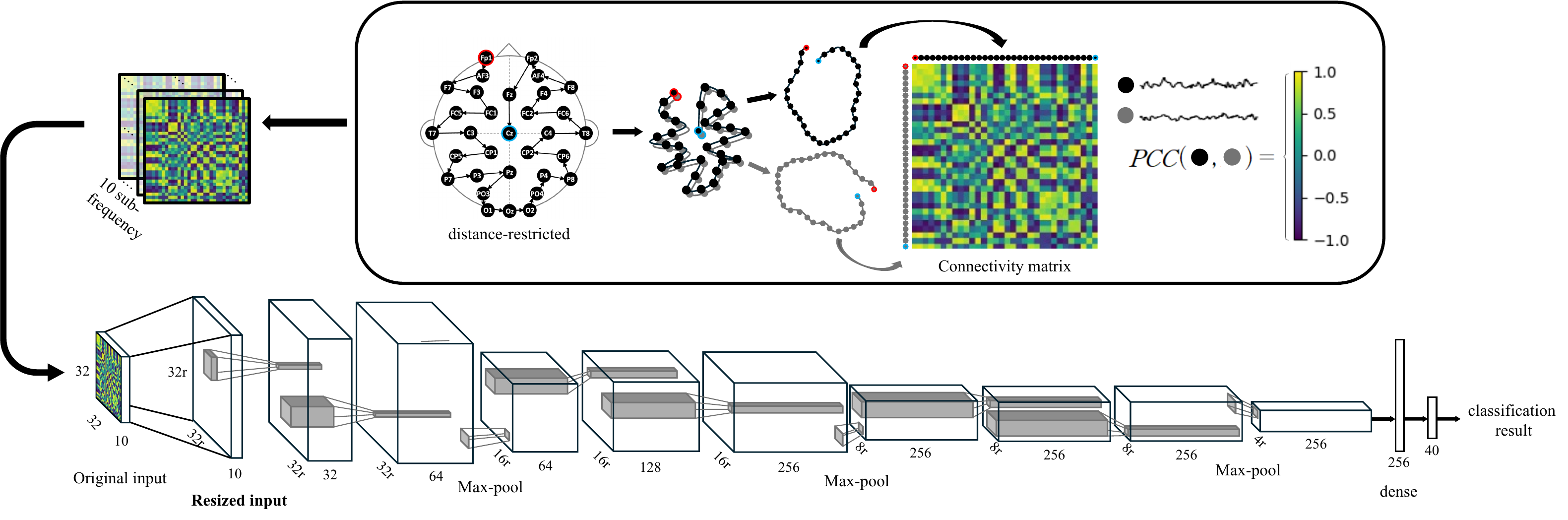}  %
    \caption{Overview of the proposed classification system}
    \label{fig:overview}
    \vspace{-0.5cm}
\end{figure*}

\subsection{Proposed Method}

To address the limitations discussed earlier, we propose a method to facilitate feature learning in CNNs by increasing the spatial resolution of connectivity matrices through interpolation techniques. This approach aims to preserve and emphasize local patterns within the connectivity matrices, ensuring that they are not lost during the convolutional operations, which ultimately leads to improved classification performance.

An overview of the proposed classification system is provided in Fig.~\ref{fig:overview}. Initially, EEG signals are preprocessed to remove noise and artifacts. The signals are then segmented into 10 distinct sub-frequency bands, as detailed in Table~\ref{table:freq}. For each sub-frequency band, a separate 32×32 connectivity matrix is generated, capturing pairwise interactions among the 32 EEG electrodes. This approach allows for the effective capture of subtle differences in neural responses across the frequency domain.

For the interpolation of the connectivity matrices, we employ the nearest-neighbor and bilinear interpolation methods. Various scaling ratios—1.5, 2, 2.5, 3, 3.5, and 4—are explored to identify the optimal resolution for feature extraction.

The upscaled connectivity matrices are then fed into the CNN model. The architecture of the proposed CNN is summarized in Table~\ref{table:model}. The model consists of successive convolutional layers with ReLU activation and max-pooling layers. The final layers include two fully connected layers, with a softmax activation function used to classify the data into 40 distinct classes (refer to Section IV for the classification task).

A key consideration when working with connectivity matrices of varying resolutions is the selection of the CNN kernel size, as it determines the receptive field within the input matrix. With changing the input resolution, a fixed kernel size corresponds to a different relative spatial extent in the input space. To evaluate this effect, we assess model performance using kernel sizes of 3×3, 5×5, and 7×7.


\begin{table}
\centering
\caption{CNN architecture in the proposed system. \\ ``$r$'' refers to the upscaling factor.} 
\begin{tabular}{|c||c|c|}
\hline
\textbf{Layer} & \textbf{Type} & \textbf{Output shape} \\
\hline
1 & Upsampling & 32r × 32r × 10 \\
2 & Convolution (ReLU) & 32r × 32r × 32 \\
3 & Convolution (ReLU) & 32r × 32r × 64 \\
4 & Max-pooling & 16r × 16r × 64 \\
5 & Convolution (ReLU) & 16r × 16r × 128 \\
6 & Convolution (ReLU) & 16r × 16r × 256 \\
7 & Max-pooling & 8r × 8r × 256 \\
8 & Convolution (ReLU) & 8r × 8r × 256 \\
9 & Convolution (ReLU) & 8r × 8r × 256 \\
10 & Max-pooling & 4r × 4r × 256 \\
11 & Dense & 256 \\
12 & Dense (Softmax) & 40 \\
\hline
\end{tabular}
\label{table:model}
\end{table}

\begin{table}
\centering
\caption{Hyperparameters for model training}
\begin{tabular}{|c|c|}
\hline
\textbf{Component} & \textbf{Setting/Value} \\
\hline
Loss function & Cross-entropy loss \\
Optimizer & Adam \\
Learning rate & 0.0001 \\
Scheduler & StepLR with stepsize 10, $\gamma$=0.8 \\
Batch size & 64 \\
Early stop sount & 20 \\
\hline
\end{tabular}
\label{table:hyper}
\end{table}

\section{Experimental Setup}

The evaluation in this study utilizes the DEAP dataset \cite{koelstra2011deap}, which contains EEG signals recorded from 32 participants while they viewed 40 emotional music videos. The EEG signals were initially recorded at a sampling rate of 512 Hz and subsequently downsampled to 128 Hz. The dataset includes emotion labels categorized by valence and arousal scales for each video stimulus. Based on the evaluation framework in \cite{moon2020emotional}, we develop a classification model to identify which of the 40 videos was being viewed based on the recorded EEG signals.

The preprocessing pipeline begins with EEG signal filtering to remove noise and artifacts. A bandpass filter is then applied to extract ten distinct frequency bands, as shown in Table~\ref{table:freq}. This granular frequency band extraction enables detailed analysis of neural responses across different spectral components of the EEG signal.

As in \cite{moon2020emotional}, the one-minute EEG signal for each trial is divided into three-second-long segments. A simple non-overlapping segmentation would yield only 20 segments per trial. To augment the dataset, we apply an overlapping segmentation approach with a step size of 0.5 seconds, thereby increasing the number of samples. For each three-second segment, a 32×32×10 connectivity matrix is computed, capturing inter-channel relationships across the ten sub-frequency bands.

From each trial, one segment is randomly selected for testing, another for validation, and the remaining segments are allocated to the training set. To prevent data leakage caused by overlapping regions, any training segments that temporally overlap with the validation or test segments are excluded. This procedure results in 119,040 segments for training, 1,280 for validation, and 1,280 for testing.

To ensure fair model evaluation, we implement a modified four-fold cross-validation strategy suited to the constraints of our overlapping segmentation scheme. Since conventional cross-validation is not applicable due to the continuous and overlapping nature of our dataset, we choose four non-overlapping test segments from each trial, each drawn from sufficiently distant temporal locations within the one-minute EEG recording to ensure exclusivity. For each test segment, the same dataset partitioning procedure is applied: one validation segment is selected among the segments non-overlapping with the test segment, and the remaining segments are used for training. This results in four distinct dataset splits that collectively form a variant of cross-validation. The final model performance is reported as the average classification accuracy over these four splits. All experiments are conducted on an NVIDIA Tesla V100 GPU.


\begin{figure}[b]
    \centering
    \includegraphics[width=0.5\textwidth]{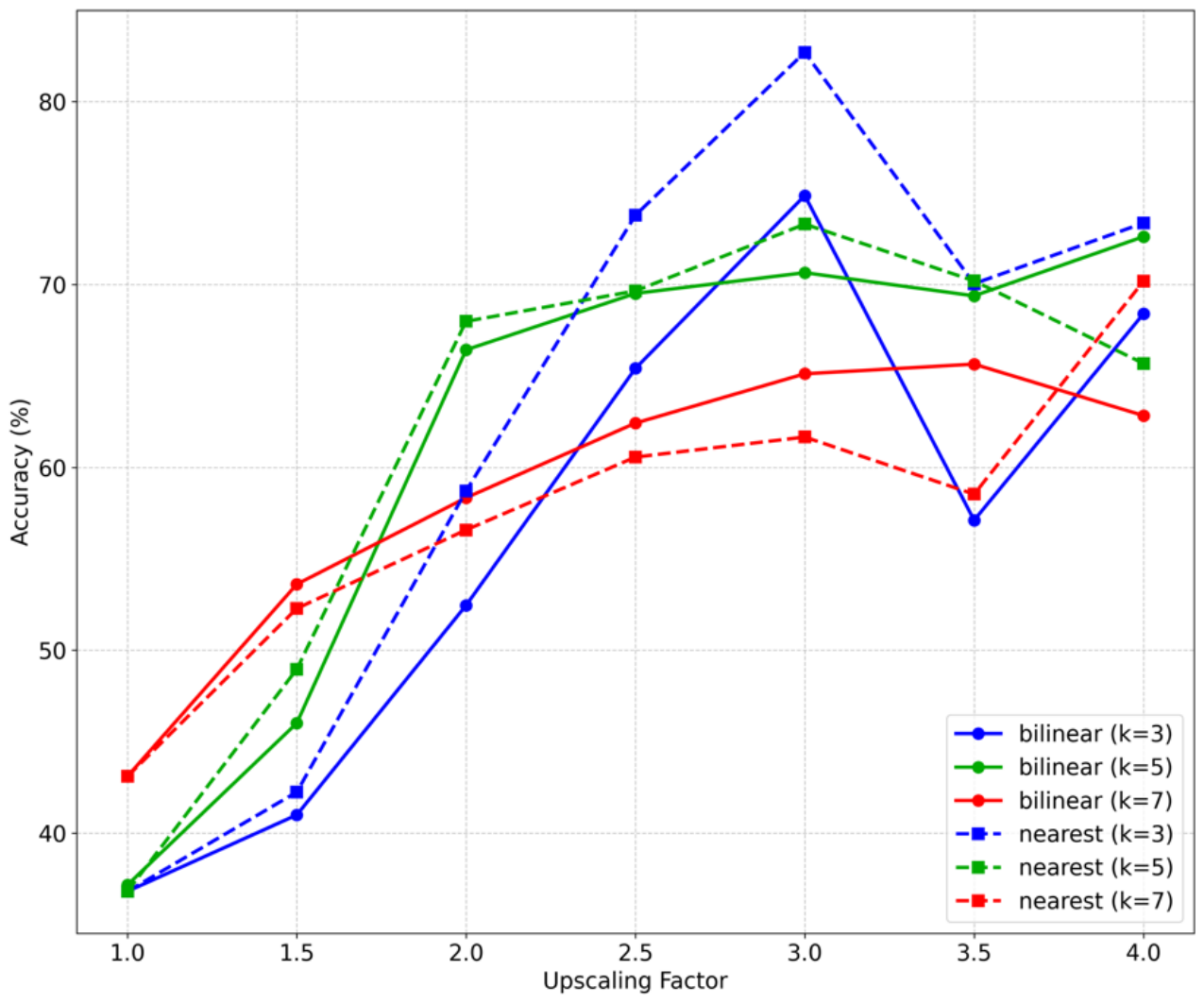}
    \caption{Classification accuracy depending on the interpolation method and kernel size across upscaling factors. The upscaling factor of 1.0 corresponds to the conventional approach.}
    \label{fig:accuracy_plot}
\end{figure}

\section{Results}
\subsection{Overall Performance}
Figure~\ref{fig:accuracy_plot} presents the classification accuracy with respect to the upscaling factor, comparing bilinear and nearest-neighbor interpolation methods across three convolutional kernel sizes ($k = 3, 5, 7$). Overall, the results confirm that upscaling connectivity matrices leads to notable improvements in classification performance across all configurations.

Among the two interpolation methods, nearest-neighbor interpolation consistently yields slightly higher accuracy than bilinear interpolation. This suggests that preserving sharp local transitions in the connectivity matrices—rather than introducing smoothed values—better supports feature extraction in CNNs, particularly in the early convolutional layers.

The impact of the kernel size on performance varies depending on the degree of upscaling. For lower upscaling factors (1.0 and 1.5), larger kernels (5×5 and 7×7) tend to outperform smaller kernels, likely due to their larger receptive fields compensating for the limited resolution. However, as the upscaling factor increases, this trend is reversed: the 3×3 kernel consistently achieves the highest accuracy at higher resolutions (i.e., ×3.0 and above). Notably, the best performance of 82.69\% is achieved using nearest-neighbor interpolation with a 3×3 kernel at an upscaling factor of 3.

These findings suggest that while larger kernels may be advantageous at low input resolutions, they provide limited benefit when fine-grained spatial details are already enhanced through upscaling. Furthermore, we observed that larger kernels sometimes exhibit training instability, particularly at high upscaling factors, occasionally resulting in failed convergence. In such cases, training was repeated until successful convergence was achieved.

These observations collectively highlight the importance of balancing the kernel size and input resolution, as well as the utility of nearest-neighbor interpolation in enhancing the spatial interpretability and classification performance of CNNs applied to EEG connectivity matrices.

\begin{figure}[b!]
    \centering
    \includegraphics[width=0.5\textwidth]{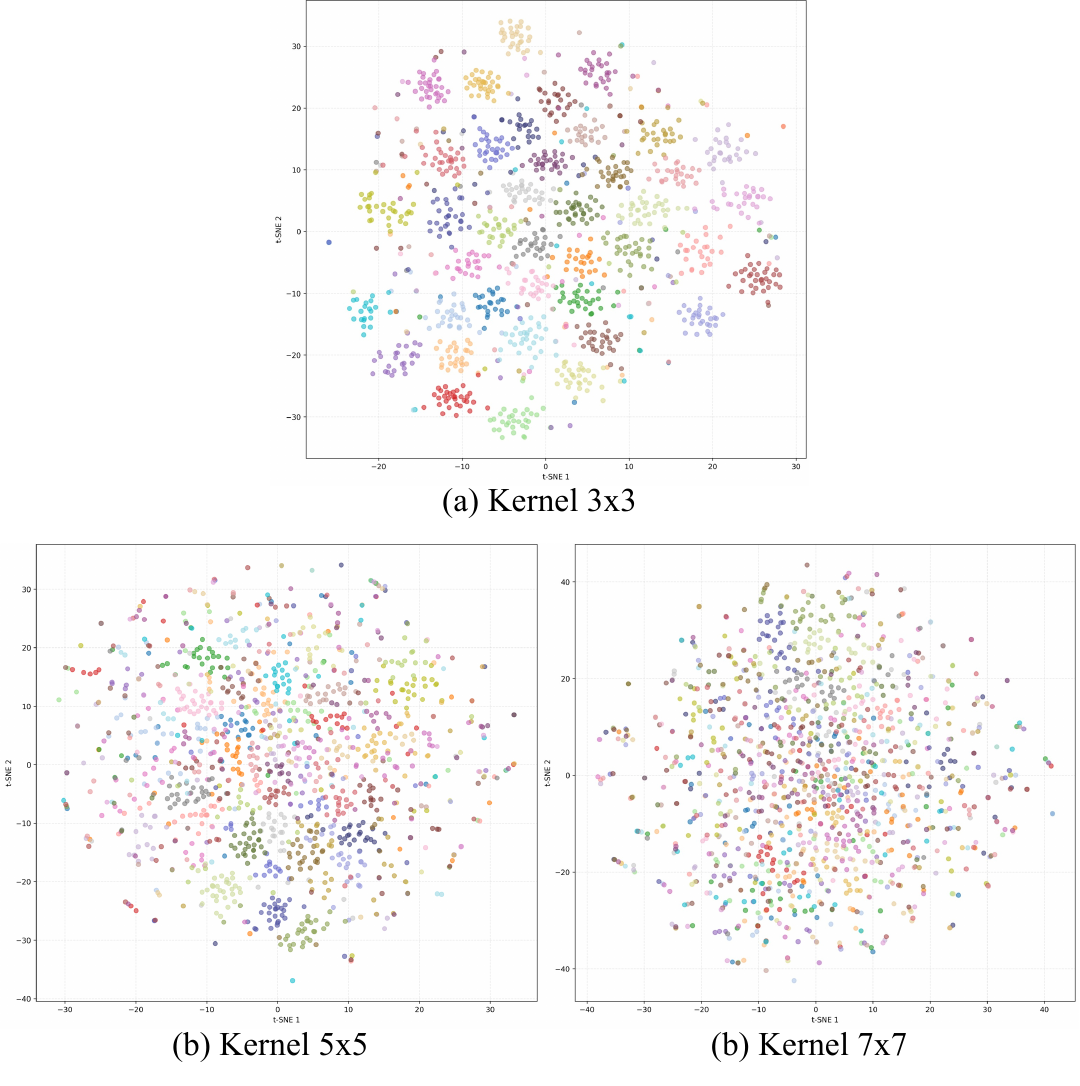}
    \caption{t-SNE visualization of the feature representations extracted by models with different convolutional kernel sizes (3×3, 5×5, 7×7) under the nearest-neighbor interpolation and 3× upscaling setting. Different colors indicate different classes.}
    \label{fig:tsne_kernels}
    \vspace{1cm}
    \centering
    \includegraphics[width=0.5\textwidth]{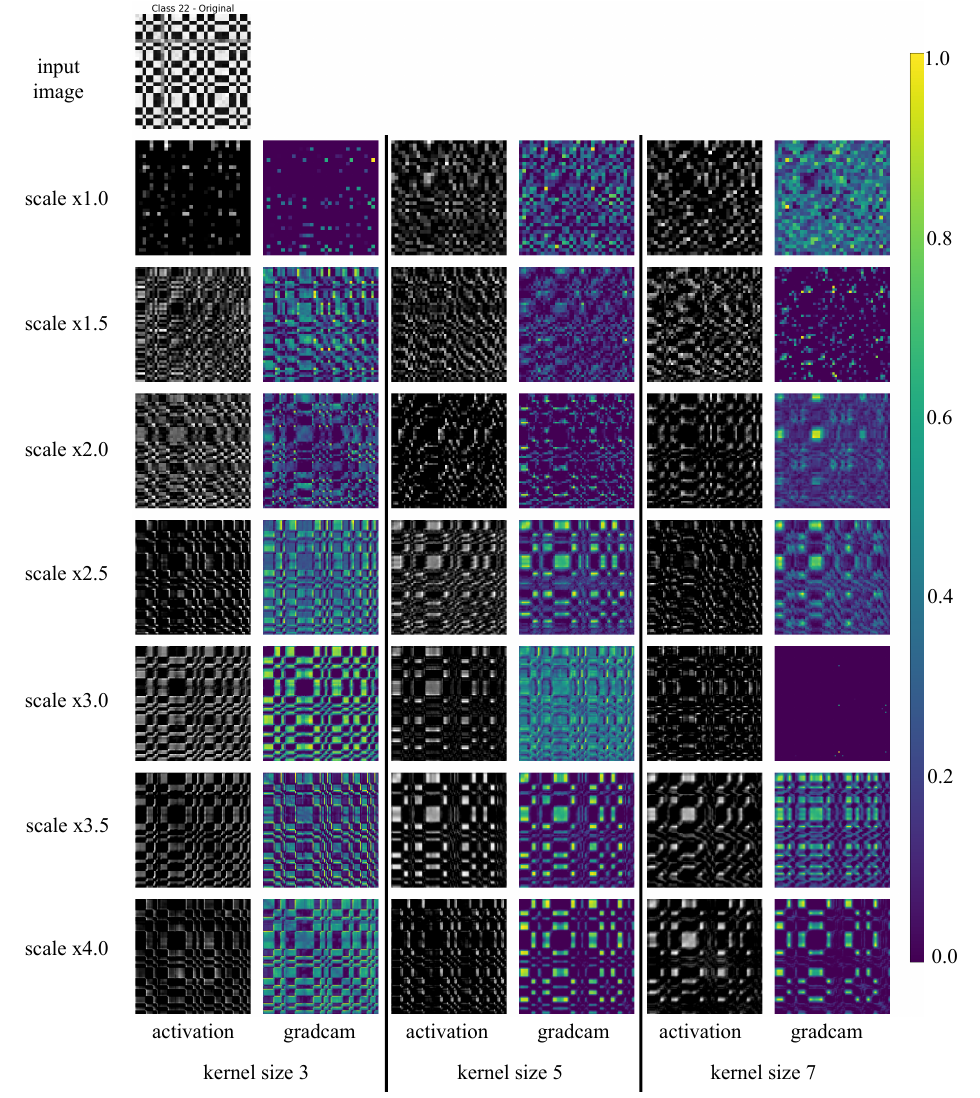}
    \caption{Activations after the first convolutional layer (left, grayscale) and corresponding Grad-CAM visualizations (right, viridis colormap), highlighting the spatial importance of early-layer features with respect to the model outputs. All results are from models using nearest-neighbor interpolation.}
    \label{fig:activ}
\end{figure} 
\subsection{Visualization of t-SNE}

To assess the discriminative quality of features extracted by the trained CNNs, we apply t-distributed stochastic neighbor embedding (t-SNE) \cite{van2008visualizing} to the final feature vectors produced using different convolutional kernel sizes ($k = 3, 5, 7$), under a fixed upscaling factor of ×3.0 with nearest-neighbor interpolation. The resulting embeddings are illustrated in Fig.~\ref{fig:tsne_kernels}.

As shown in Fig.~\ref{fig:tsne_kernels}(a), when using the 3×3 kernel, the feature embeddings form compact and well-separated clusters, indicating strong class-specific representations. Each class occupies a distinct region in the embedded space, suggesting that the model effectively captures the underlying differences in EEG connectivity patterns.

When the kernel size increases to 5×5 (Fig.~\ref{fig:tsne_kernels}(b)), cluster separability degrades. Although some class distinction remains, the embeddings are more diffused, and there is noticeable overlap between classes.

For the 7×7 kernel (Fig.~\ref{fig:tsne_kernels}(c)), the feature space becomes heavily entangled with substantial overlap across all classes. The lack of clear separation implies that large kernels may oversmooth fine spatial details, ultimately degrading the model's ability to learn distinct representations.

\begin{figure*}[h!]
    \centering
    \includegraphics[width=0.6\textwidth]{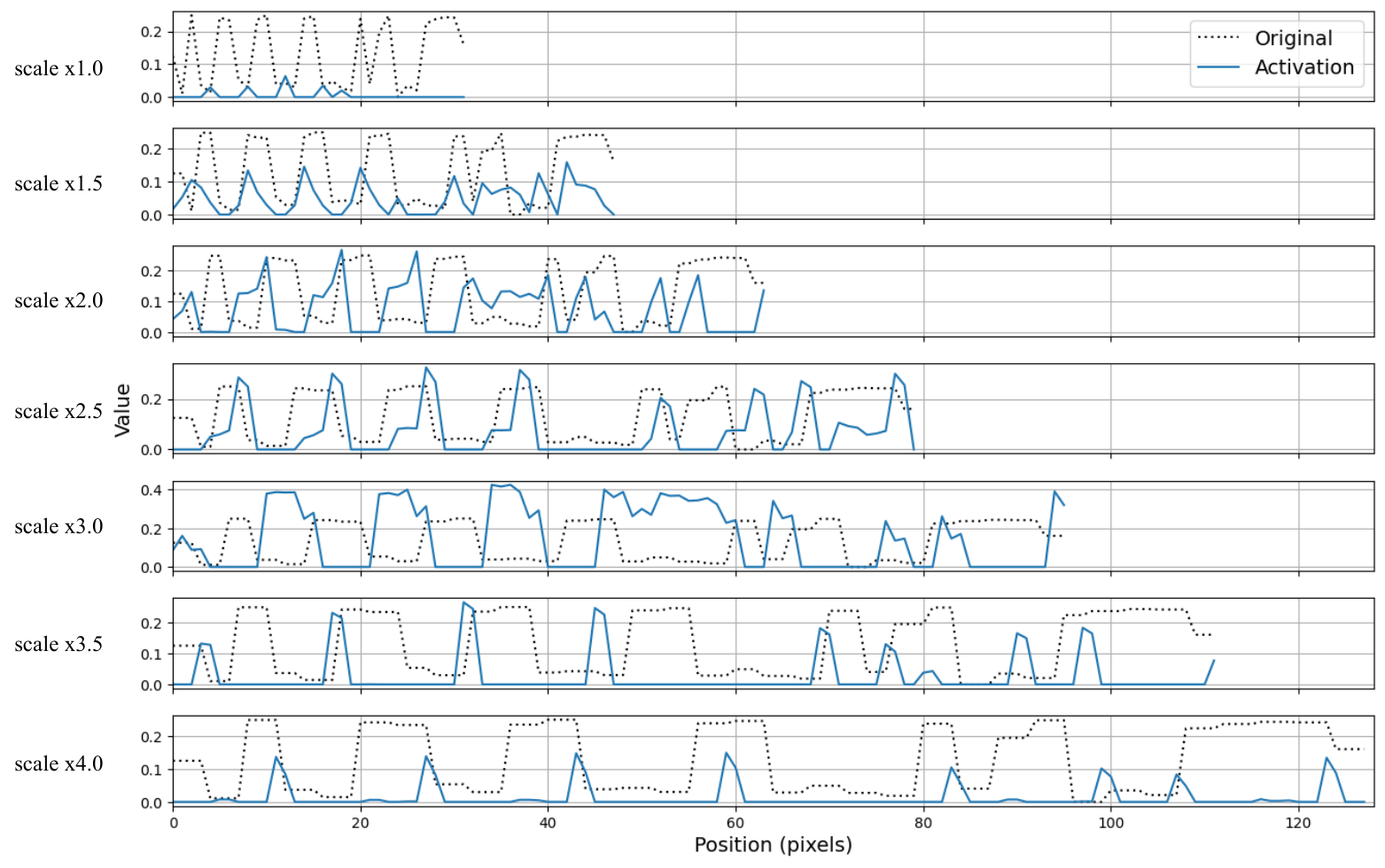}
    \caption{One-dimensional plots of the input and the activations of the first convolutional layer along a particular row when the kernel size is 3×3.}
    \label{fig:all_line}
    \vspace{-0.1cm}
\end{figure*}

\subsection{Visualization of CNN Activation}

To further understand how the input resolution affects feature learning, we visualize the activation maps from the first convolutional layer along with their corresponding Grad-CAM~\cite{selvaraju2017grad} outputs, which highlight the regions contributing most to the model’s final prediction. As illustrated in Fig.~\ref{fig:activ}, higher upscaling factors lead to more spatially coherent and sharply defined activation patterns. This observation suggests that increased input resolution allows the model to more effectively exploit the spatial structure inherent in the connectivity matrices.

Grad-CAM maps further validate this effect, showing that the refined activation regions at higher resolutions align with semantically relevant parts of the input. This implies that the model, even in its early layers, begins to focus on areas that are crucial for accurate classification. Notably, the 3×3 kernel at an upscaling factor of ×3.0, which shows the highest accuracy in Fig. \ref{fig:accuracy_plot}, yields the clearest Grad-CAM results, demonstrating the model’s ability to effectively focus on the most relevant features for classification.

In contrast, smaller upscaling factors exhibit clear loss of spatial structures in the Grad-CAM outputs, showing less coherent and more diffuse activation patterns. Particularly, at ×1.0 resolution, the Grad-CAM results show complete loss of the original pattern, with no recognizable structure remaining in the activation maps. This suggests that the model is unable to properly capture the fine spatial relationships within the connectivity matrix at this resolution, severely hindering its ability to identify discriminative features.

To gain more detailed understanding of whether higher-resolution inputs indeed lead to more informative activations being passed to the model, we examine one-dimensional activation patterns along a representative row of the connectivity matrix, as shown in Fig.~\ref{fig:all_line}. At the native resolution (×1.0 scaling), the activations are relatively sparse, with only five small peaks activated, limitedly capturing local structures.

As the scaling factor increases up to ×3.0, activations become rich and well-distributed across the input, leading to the enhanced classification accuracy as shown in Fig. \ref{fig:accuracy_plot}. However, at ×3.5 and beyond, the expended pixels result in excessive spatial redundancy, making it difficult for the model to focus on essential features. Consequently, activations become confined to pixel boundaries and it becomes hard for the model to capture the underlying meaningful connectivity structures.

\newpage
\section{Conclusion}

This paper has presented a method for improving emotional EEG classification using CNNs by upscaling input connectivity matrices. We identified the limitations of using original connectivity matrices in their native resolution, i.e., applying CNNs to connectivity matrices often fail to capture the relevant patterns when compared to natural images. To address this, we suggested to employ an interpolation technique that preserves localized structures in the connectivity matrices. Our experimental results demonstrated significant improvement in classification accuracy by our method. Higher resolution inputs (specifically ×3.0 scaling) provided more informative activations, leading to improved model performance. Overall, our interpolation-based enhancement strategy successfully improved EEG connectivity-based emotion recognition. In our future work, we plan to extend our work to develop adaptive resolution techniques or hybrid models to further enhance classification performance.

\section*{Acknowledgement}
This work was supported by the National Research Foundation of Korea (NRF) grant funded by the Korea government (MSIT) (No. RS-2024-00453301) and by Institute of Information \& communications Technology Planning \& Evaluation (IITP) under 6G Cloud Research and Education Open Hub (IITP-2025-RS-2024-00428780) grant funded by the Korea government (MSIT). This work was also supported by the Yonsei Signature Research Cluster Program of 2024 (2024-22-0161).

\bibliographystyle{unsrt}
\bibliography{references}
\end{document}